\documentclass[a4paper,11pt]{article}

\usepackage{breakurl}
\usepackage[breaklinks=true]{hyperref}
\usepackage{breakcites}

\usepackage{amssymb,amsmath,array,bm}
\usepackage{booktabs}

\usepackage{tikz}
\usetikzlibrary{arrows}

\usepackage{pgfplots}
\pgfplotsset{compat=1.15}
\usetikzlibrary{arrows}

\usepackage[a4paper, left=3cm, right=2.5cm, top=3cm, bottom=3cm]{geometry}

\usepackage{authblk}
\usepackage{natbib}
\usepackage{doi}

\usepackage[utf8]{inputenc}
\usepackage[T1]{fontenc}
\usepackage[english]{babel}
\usepackage{csquotes}

\title{Evaluation of LLM-based Explanations for a Learning Analytics Dashboard}

\author[1]{Alina Deriyeva}
\author[1]{Benjamin Paassen}
\affil[1]{Faculty of Technology, Bielefeld University}

\date{preprint as provided by the authors}

\begin{document}

\maketitle

\pagestyle{myheadings}
\markright{preprint as provided by the authors}

\begin{abstract}
Learning Analytics Dashboards can be a powerful tool to support self-regulated learning in Digital Learning Environments and promote development of meta-cognitive skills, such as reflection. However, their effectiveness can be affected by the interpretability of the data they provide. To assist in the interpretation, we employ a large language model to generate verbal explanations of the data in the dashboard and evaluate it against a standalone dashboard and explanations provided by human teachers in an expert study with university level educators (N=12). We find that the LLM-based explanations of the skill state presented in the dashboard, as well as general recommendations on how to proceed with learning within the course are significantly more favored compared to the other conditions. This indicates that using LLMs for interpretation purposes can enhance the learning experience for learners while maintaining the pedagogical standards approved by teachers. 
\end{abstract}

\section{Introduction}

With the raise of the demand in education, Digital Leaning Environments (DLEs) and particularly Intelligent Tutoring Systems (ITSs) can serve as powerful tools to bridge the gap between demand and supply as well as enrich the overall learning experience \cite{deriyeva2025}. For example, ITSs have been shown to be efficient to support both self-regulated and teacher-guided learning \cite{kulik_2016}. To promote meta-cognitive skills and support self-regulation of learning processes, many DLEs are using dashboards to show learning analytics on student’s ability development and learning behavior to help with reflection on the progress and support decision-making. However, in many cases, additional explanations are needed to make sense of the information presented in a dashboard and act on it appropriately \cite{matcha2020}. To evaluate perceived effectiveness of such explanations, we conduct a study with human experts (N=12) to evaluate  different types of explanations for an Open Learner Model dashboard \cite{bodily2018} presenting data from a Performance Factors Analysis (PFA) model \cite{pavlik2009}. We find that, on average, teachers tend to favor LLM-based explanations and recommendations over those  provided by a human teacher. The data collected as well as the respective prompts and analysis code are available in a git repository \footnote{\href{https://gitlab.ub.uni-bielefeld.de/publications-ag-kml/xlm-explanations-evaluation}{https://gitlab.ub.uni-bielefeld.de/publications-ag-kml/xlm-explanations-evaluation}}. 

\section{Case Study}

\begin{figure}
    \centering
    \includegraphics[width = 15.5cm]{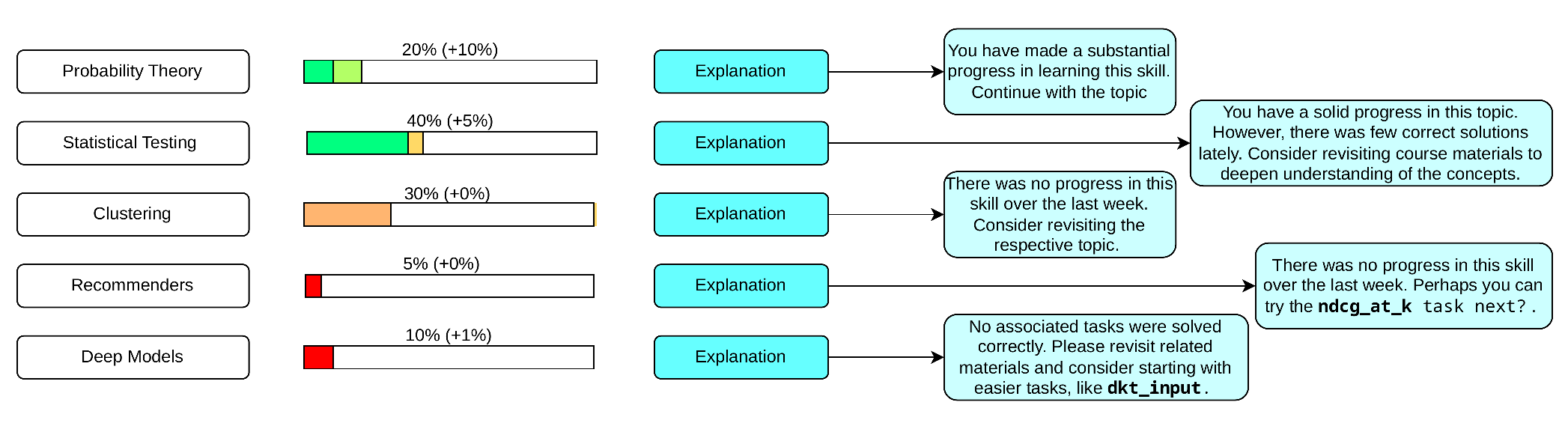}
    \caption{Example of the skill state explanations within the sketch of the dashboard (Condition B).}
    \label{fig:skill_example}
\end{figure}

To investigate the perceived effectiveness of the explanations of the dashboard, we conducted a within-subject study with N=12 participants, where every participant was exposed to 3 conditions in a randomized order: A) a dashboard sketch with no additional explanations, B) a dashboard enriched with explanations and recommendations provided by human teachers (cross-checked by 2 teachers), and C) a dashboard enriched with explanations and recommendations provided by a large language model (Qwen 32B, provided through BIKI interface\footnote{\href{https://www.uni-bielefeld.de/einrichtungen/bits/services/kuz/biki/}{https://www.uni-bielefeld.de/einrichtungen/bits/services/kuz/biki/}}, prompts can be found in the git repository).  For each condition, two simulated example students were presented with different learning performance. After each condition, the participants were asked to evaluate the information they were shown with 3 questionnaires: TOAST \cite{wojton2020} and two adjusted explanation satisfaction scales \cite{hoffman2023}, one for the explanation of the PFA-based skill states and one for the general course recommendation. Figure \ref{fig:skill_example} shows one of the  examples from condition B. 

Additionally, the participants were asked to fill in pre- and post-questionnaires to assess their level of experience with teaching as well as strategies they tend to employ to promote the self-regulated learning of the students and to give feedback, as well as to collect their feedback and opinions regarding the study and potential ways to improve the system and explanations they were presented with. The recruited participants were people with university-level teaching experience. The study was approved by the local ethics committee. 

\section{Results}

After the data collection, the questionnaire evaluation results for different conditions were compared against each other with pair-wise t-tests. Firstly, the results of the TOAST evaluation of the three conditions show that there is no little to no difference in the Understanding factor (refers to the understanding of what system needs to do). However, the evaluation of the performance is significantly (p << 0.5) better for the LLM condition, with Cohen’s d of 0.96 compared to the condition with explanations from a human teacher.  A similar pattern is reflected in Table \ref{tab:results} showing that LLM-generated explanations to the skill states (right) and recommendations on how to proceed within the course (left) are heavily favored over the two other conditions.  While the study is subject to several limitations (sample size, homogeneous population, imperfections in teacher-generated feedback) and the topic would benefit from a more extensive investigation, these results suggest that more extensive and detailed explanations and recommendations generated by LLMs are considered surprisingly effective by teachers. Hence, in our specific case, we conclude that the LLM can indeed be used to enhance students’ learning experience via automatic explanations of learning analytic dashboards.

\begin{table}[]
    \centering
    \begin{tabular}{|c|c|c|c|c|}
        \hline
         & \multicolumn{2}{|c|}{Recommendations} & \multicolumn{2}{|c|}{Skill State Explanations} \\ \hline
         & Condition B & Condition C & Condition B & Condition C \\ \hline
        Condition A & d = -2.4 & d = -3.65 & d = -0.82 & d = -2.20 \\ \hline
        Condition B & x & d = -0.95 & x & d = -1.34 \\ \hline

    \end{tabular}
    \caption{Effect size (Cohen's d) in the pairwise comparison of the evaluation of the conditions. Negative sign indicates preference towards column headers. All the results presented for p << 0.01.}
    \label{tab:results}
\end{table}
\section*{Acknowledgement}

This work has been part of the project “Explaining Learner Models using Language Models” as part of KI:edu.nrw. KI:edu.nrw is funded by the Ministry of Culture and Science of the State of North Rhine-Westphalia. We gratefully
acknowledge this support.

\bibliographystyle{plainnat}
\bibliography{LAK_Poster}

\begin{thebibliography}{7}
\providecommand{\natexlab}[1]{#1}
\providecommand{\url}[1]{\texttt{#1}}
\expandafter\ifx\csname urlstyle\endcsname\relax
  \providecommand{\doi}[1]{doi: #1}\else
  \providecommand{\doi}{doi: \begingroup \urlstyle{rm}\Url}\fi

\bibitem[Bodily et~al.(2018)Bodily, Kay, Aleven, Jivet, Davis, Xhakaj, and
  Verbert]{bodily2018}
Robert Bodily, Judy Kay, Vincent Aleven, Ioana Jivet, Dan Davis, Franceska
  Xhakaj, and Katrien Verbert.
\newblock Open learner models and learning analytics dashboards: a systematic
  review.
\newblock In \emph{Proceedings of the 8th {International} {Conference} on
  {Learning} {Analytics} and {Knowledge}}, pages 41--50, Sydney New South Wales
  Australia, March 2018. ACM.
\newblock ISBN 978-1-4503-6400-3.
\newblock \doi{10.1145/3170358.3170409}.
\newblock URL \url{https://dl.acm.org/doi/10.1145/3170358.3170409}.

\bibitem[Deriyeva et~al.(2025)Deriyeva, Dannath, and Paassen]{deriyeva2025}
Alina Deriyeva, Jesper Dannath, and Benjamin Paassen.
\newblock {SCRIPT}: {Implementing} an {Intelligent} {Tutoring} {System} for
  {Programming} in a {German} {University} {Context}.
\newblock In Alexandra~I. Cristea, Erin Walker, Yu~Lu, Olga~C. Santos, and
  Seiji Isotani, editors, \emph{Artificial {Intelligence} in {Education}.
  {Posters} and {Late} {Breaking} {Results}, {Workshops} and {Tutorials},
  {Industry} and {Innovation} {Tracks}, {Practitioners}, {Doctoral}
  {Consortium}, {Blue} {Sky}, and {WideAIED}}, volume 2590, pages 168--177.
  Springer Nature Switzerland, Cham, 2025.
\newblock ISBN 978-3-031-99260-5 978-3-031-99261-2.
\newblock \doi{10.1007/978-3-031-99261-2_16}.
\newblock URL \url{https://link.springer.com/10.1007/978-3-031-99261-2_16}.
\newblock Series Title: Communications in Computer and Information Science.

\bibitem[Hoffman et~al.(2023)Hoffman, Mueller, Klein, and Litman]{hoffman2023}
Robert~R. Hoffman, Shane~T. Mueller, Gary Klein, and Jordan Litman.
\newblock Measures for explainable {AI}: {Explanation} goodness, user
  satisfaction, mental models, curiosity, trust, and human-{AI} performance.
\newblock \emph{Frontiers in Computer Science}, 5:\penalty0 1096257, February
  2023.
\newblock ISSN 2624-9898.
\newblock \doi{10.3389/fcomp.2023.1096257}.
\newblock URL
  \url{https://www.frontiersin.org/articles/10.3389/fcomp.2023.1096257/full}.

\bibitem[Kulik and Fletcher(2016)]{kulik_2016}
James~A. Kulik and J.~D. Fletcher.
\newblock Effectiveness of {Intelligent} {Tutoring} {Systems}: {A}
  {Meta}-{Analytic} {Review}.
\newblock \emph{Review of Educational Research}, 86\penalty0 (1):\penalty0
  42--78, March 2016.
\newblock ISSN 0034-6543, 1935-1046.
\newblock \doi{10.3102/0034654315581420}.
\newblock URL \url{https://journals.sagepub.com/doi/10.3102/0034654315581420}.

\bibitem[Matcha et~al.(2020)Matcha, Uzir, Gasevic, and Pardo]{matcha2020}
Wannisa Matcha, Noraayu~Ahmad Uzir, Dragan Gasevic, and Abelardo Pardo.
\newblock A {Systematic} {Review} of {Empirical} {Studies} on {Learning}
  {Analytics} {Dashboards}: {A} {Self}-{Regulated} {Learning} {Perspective}.
\newblock \emph{IEEE Transactions on Learning Technologies}, 13\penalty0
  (2):\penalty0 226--245, April 2020.
\newblock ISSN 1939-1382, 2372-0050.
\newblock \doi{10.1109/TLT.2019.2916802}.
\newblock URL \url{https://ieeexplore.ieee.org/document/8713912/}.

\bibitem[{Pavlik Philip I.} et~al.(2009){Pavlik Philip I.}, {Cen Hao}, and
  {Koedinger Kenneth R.}]{pavlik2009}
{Pavlik Philip I.}, {Cen Hao}, and {Koedinger Kenneth R.}
\newblock Performance {Factors} {Analysis} \&ndash; {A} {New} {Alternative} to
  {Knowledge} {Tracing}.
\newblock In \emph{Frontiers in {Artificial} {Intelligence} and
  {Applications}}. IOS Press, 2009.
\newblock \doi{10.3233/978-1-60750-028-5-531}.
\newblock URL
  \url{https://www.medra.org/servlet/aliasResolver?alias=iospressISSNISBN&issn=0922-6389&volume=200&spage=531}.

\bibitem[Wojton et~al.(2020)Wojton, Porter, T.~Lane, Bieber, and
  Madhavan]{wojton2020}
Heather~M. Wojton, Daniel Porter, Stephanie T.~Lane, Chad Bieber, and Poornima
  Madhavan.
\newblock Initial validation of the trust of automated systems test ({TOAST}).
\newblock \emph{The Journal of Social Psychology}, 160\penalty0 (6):\penalty0
  735--750, November 2020.
\newblock ISSN 0022-4545, 1940-1183.
\newblock \doi{10.1080/00224545.2020.1749020}.
\newblock URL
  \url{https://www.tandfonline.com/doi/full/10.1080/00224545.2020.1749020}.

\end{thebibliography}
\end{document}